\documentclass[journal]{IEEEtran}
\usepackage{booktabs}
\usepackage{subfigure}
\usepackage{graphicx}
\usepackage{float}
\usepackage{graphicx,amssymb,mathrsfs,amsmath,latexsym}
\usepackage{epsfig}
\usepackage{epstopdf}
\usepackage{xcolor}
\usepackage{algorithm}
\usepackage{algpseudocode}
\usepackage{cite}
\usepackage{mathrsfs}
\usepackage[colorlinks,linkcolor=black,anchorcolor=black,citecolor=black]{hyperref}
\usepackage{multirow}
\usepackage{array}
\usepackage[switch]{lineno}
\newcommand{\PreserveBackslash}[1]{\let\temp=\\#1\let\\=\temp}
\newcolumntype{C}[1]{>{\PreserveBackslash\centering}p{#1}}
\newcolumntype{R}[1]{>{\PreserveBackslash\raggedleft}p{#1}}
\newcolumntype{L}[1]{>{\PreserveBackslash\raggedright}p{#1}}

\makeatletter

\begin{document}
\title{Fourier-based Rotation-invariant Feature Boosting: An Efficient Framework for Geospatial Object Detection}

\author{Xin~Wu,~\IEEEmembership{Student Member,~IEEE,}
        Danfeng Hong,~\IEEEmembership{Student Member,~IEEE,}
        Jocelyn Chanussot,~\IEEEmembership{Fellow,~IEEE,}
        Yang Xu,~\IEEEmembership{Member,~IEEE,}
        Ran Tao,~\IEEEmembership{Senior Member,~IEEE,}
        Yue Wang
\thanks{This work was supported in part by the National Natural Science Foundation of China under Grant 61421001 and Grant U1833203, as well as supported in part by the National Natural Science Foundation of China under  Grant 61701238, in part by the Jiangsu Provincial Natural Science Foundation of China under Grant BK20170858. (\emph{Corresponding author: Danfeng Hong.})}
\thanks{X. Wu, R. Tao and Y. Wang are with the School of Information and Electronics, Beijing Institute of Technology 100081, China, and Beijing Key Laboratory of Fractional Signals and Systems, School of Information and Electronics, Beijing Institute of Technology, Beijing 100081, China. (e-mail: hdfwx@bit.edu.cn).}
\thanks{D. Hong is with the Remote Sensing Technology Institute (IMF), German Aerospace Center (DLR), 82234 Wessling, Germany, and Signal Processing in Earth Observation (SiPEO), Technical University of Munich (TUM), 80333 Munich, Germany. (e-mail: danfeng.hong@dlr.de)}
\thanks{J. Chanussot is with the Univ. Grenoble Alpes, CNRS, Grenoble INP, GIPSA-lab, F-38000 Grenoble, France, also with the Faculty of Electrical and Computer Engineering, University of Iceland, Reykjavik 101, Iceland. (e-mail: jocelyn@hi.is)}
\thanks{Y. Xu is with the School of Computer Science and Engineering, Nanjing University of Science and Technology, Nanjing 210094, China. (e-mail: xuyangth90@njust.edu.cn).}
}

\markboth{Submission to IEEE Geoscience and Remote Sensing Letters,~Vol.~XX, No.~XX,~2019}%
{Shell \MakeLowercase{\textit{et al.}}: Bare Demo of IEEEtran.cls for IEEE Journals}

\maketitle

\begin{abstract}
\textcolor{blue}{This is the pre-acceptance version, to read the final version please go to IEEE Geoscience and Remote Sensing Letters on IEEE Xplore.} Geospatial object detection of remote sensing imagery has been attracting an increasing interest in recent years, due to the rapid development in spaceborne imaging. Most of previously proposed object detectors are very sensitive to object deformations, such as scaling and rotation. To this end, we propose a novel and efficient framework for geospatial object detection in this letter, called Fourier-based rotation-invariant feature boosting (FRIFB). A Fourier-based rotation-invariant feature is first generated in polar coordinate. Then, the extracted features can be further structurally refined using aggregate channel features. This leads to a faster feature computation and more robust feature representation, which is good fitting for the coming boosting learning. Finally, in the test phase, we achieve a fast pyramid feature extraction by estimating a scale factor instead of directly collecting all features from image pyramid. Extensive experiments are conducted on two subsets of NWPU VHR-10 dataset, demonstrating the superiority and effectiveness of the FRIFB compared to previous state-of-the-art methods.
\end{abstract}


\IEEEpeerreviewmaketitle
\section{Introduction}
\IEEEPARstart{R}{ecently}, geospatial object detection (GOD) received a lot of attention in remote sensing community. However, the main challenges and difficulties lie in that
objects in optical remote sensing imagery usually suffer from various deformations caused by scaling, offset, and rotation. This inevitably degrades the detection performances. Regarding this issue, related work has been largely proposed by researchers over the past decades. They can be roughly categorized by \emph{template matching-based, knowledge-based, object-based, and machine learning-based} methods \cite{cheng2016survey}. But unfortunately, these approaches mostly fail to capture rotation-related properties under situations of small-scale training samples.

The multi-resolution object rotation is a common but challenging problem in the task of GOD, which can be split into two sub-problems: rotation-invariant feature extraction and image pyramid generation, respectively. In the first phase, the features can be learned from the data \cite{hong2018joint,Cheng2016} or artificially designed \cite{hong2015novel,wu2019orsim}. The former learns a robust and discriminative feature representation from the augmented training set generated by manually rotating or shifting samples, whose performance is limited by the quantity and diversity of samples to a great extent, while the latter extracts the rotation-invariant features in a densely sampling fashion. Although such scheme of manual feature design has been proven to be effective (e.g., Histogram of Oriented Gradients (HOG) \cite{hong2016robust}) in constructing rotation-invariant descriptors, yet the expensive computational cost and time-consuming nature hinder it from being efficient, particularly for large-scale datasets. Moreover, those artificial descriptors also yield a relatively limited performance, since they are usually constructed in a locally discrete coordinate system. For this reason, Liu \textit{et al.} mathematically proved the rotation-invariant behavior and proposed a FourierHOG descriptor by converting a discrete coordinate system to a continuous one in \cite{IJCV2014}, where they applied the features to address a recognition-like detection problem, that is, each pixel or sub-pixel is represented as an object or material, and thus this is actually a pixel-wise classification issue rather than a real object detection one. In the second phase, the features have to be repeatedly extracted from each layer of image pyramid, leading to a large computational cost. Facing this problem, inspired by fractal statistics of natural images, Dollar \textit{et al.} \cite{PAMI2014} proposed a fast pyramid generative model (FPGM) by only estimating a scale factor, basically achieving a pyramid feature extraction in parallel.
\begin{figure*}[!t]
\centering\includegraphics[width=0.9\textwidth]{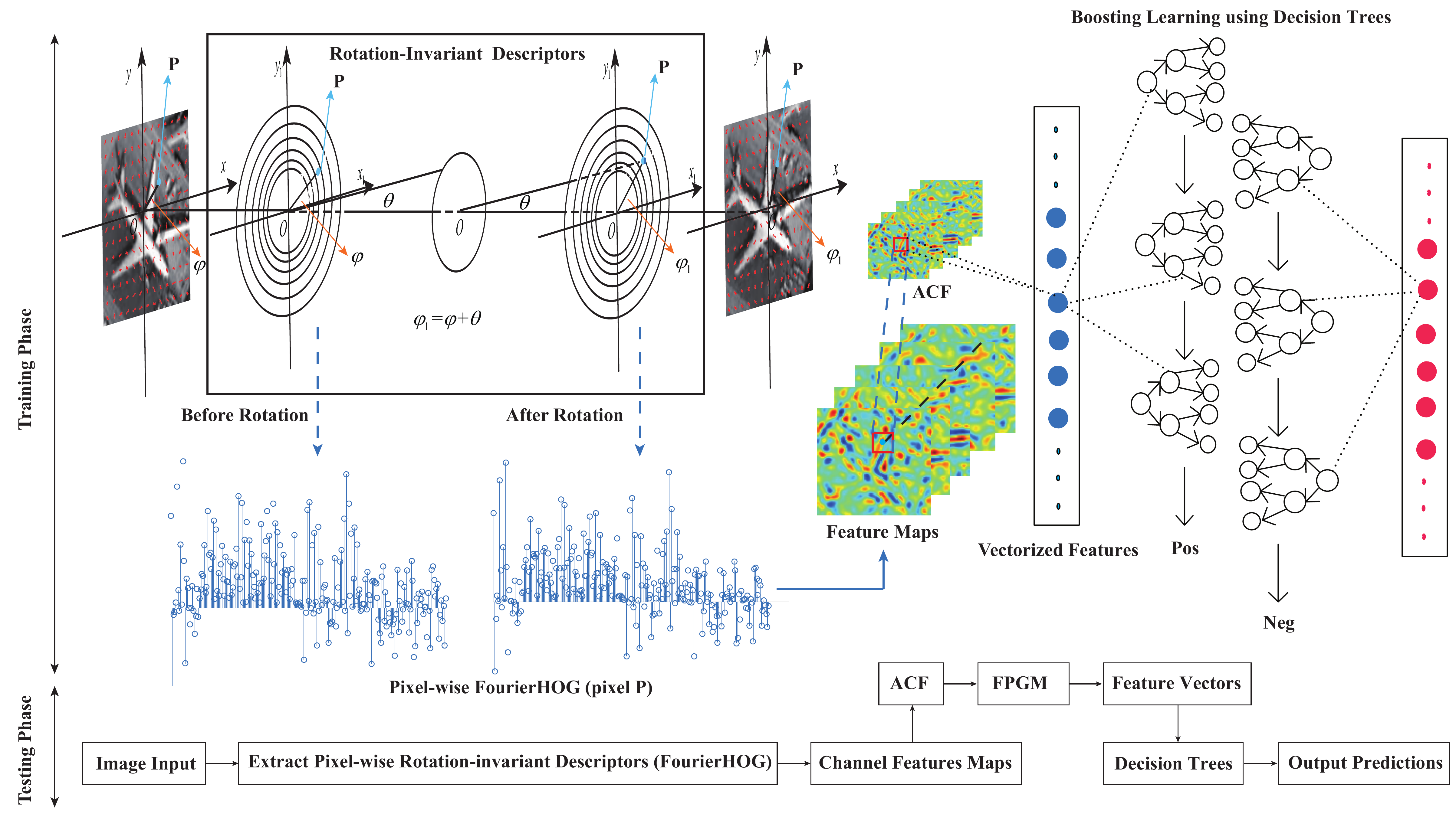}
\caption{ The workflow of the proposed FRIFB in the process of training and test.}
\label{fig1}
\end{figure*}
\subsection{Motivation}
Object rotation in GOD is an important factor to degrade the detection performance. Most previously-proposed methods usually fail to extract the continuous rotation-invariant features, since either manual feature extraction \cite{hong2016robust} or deep learning-based strategy \cite{wu2018msri,yang2018automatic} models the rotation behaviors in the discrete coordinate, such as, dividing the angles into several discrete bins or rotating the training samples with different angles for data augmentation.

On the other hand, FPGM has been proven to be effective to achieve a very fast pyramid feature extraction without the additional performance loss \cite{PAMI2014}. It should be noted that FPGM has to meet a low-level shift-invariant input, they are RGB, gray, and gradient channels used in the original reference \cite{PAMI2014}. However, these features are relatively poor discriminative and sensitive to the object rotation. Therefore, we expect to develop a more discriminative and robust feature descriptor and embed it into FPGM.

Motivated by the aforementioned two points, we expect to develop or find a mathematically rotation-invariant descriptor (FourierHOG in our case) against rotation behaviors of arbitrary continuous angle. In the meantime, the FourierHOG can be embedded into FPGM well with the requirement of low-level shift-invariance, in order to achieve an effective object detection framework.

\subsection{Contributions}
For this purpose, we propose a novel geospatial object detection framework by effectively integrating FourierHOG channel features, aggregate channel features (ACF) \cite{PAMI2014}, FPGM, and boosting learning. To the best of our knowledge, this is the first time that FPGM and boosting learning have been jointly applied to a unified geospatial object detection framework. With the further FourierHOG and ACF embedding, we have demonstrated the superiority and effectiveness using the proposed FRIFB detector on two subsets (baseballs and airplanes) of NWPU VHR-10 dataset. More specifically, the main contributions of this letter can be unfolded as
\begin{itemize}
\item An efficient geospatial object detection framework is proposed, called Fourier-based rotation-invariant feature boosting (FRIFB), encompassing feature extraction (rotation-invariant FourierHOG), feature refining (ACF), feature pyramid (FPGM), and boosting learning (decision tree ensembles).
\item The complementary advantages between FPGM and FourierHOG improve the performance of object detection in a fast and robust fashion. The robustness of the FourierHOG against rotation and shift, on one hand, perfectly fits the assumpation of the FPGM; on the other hand, FPGM can provide a faster pyramid feature computation.
\item The proposed FRIFB is effectively applied for the task of geospatial object detection in remote sensing imagery and meanwhile qualitatively and quantitatively evaluated on two different datasets and shows competitive performances against previous state-of-art algorithms.
\end{itemize}
\section{Methodology}
\subsection{Overview}
Fig. \ref{fig1} illustrates the workflow of the FRIFB, which consists of main five steps: set the sliding window, generate rotation-invariant channel features, refine channel features, training with multiple rounds of bootstrapping, and testing on image pyramid with octave-paced scale intervals. Step by step,
\begin{itemize}
\item we first give a fixed bounding box for all training samples. Generally, the size of bounding box is assigned by averaging all training samples. Targets in the different scale spaces are accordingly upsampled or downsampled to the same size.
\item Next, the corresponding rotation-invariant channel maps are obtained using the FourierHOG algorithm.
\item The ACF is subsequently used to structurally refine the extracted rotation-invariant channel features.
\item The obtained ACF is further fed into the bootstrapping for training.
\item Finally, FPGM is explored to fast generate feature pyramid features during the testing process.
\end{itemize}

\textbf{Algorithm 1} details the specific procedures for the FRIFB.
\subsection{Rotation-invariant Feature Generation (FourierHOG)}
Superior to the vector-valued function, the scalar-valued function is invariant to rotation or shift behavior. Given an image $\mathbf{I}(x,y)$: $x,y \to \mathbf{I}(x,y)$, $(x,y)$ denotes the location of a given pixel. The rotation of scalar-valued function is a coordinate transform rotation $T_g$ \cite{IJCV2014}, we have
\begin{equation}
{\mathbf{I}_{rot}}(x,y): = \mathbf{I}\left({{T_g}\left( x,y \right)} \right) = \mathbf{I}\left( {\mathbf{R}_g^{ - 1}(x,y)} \right) = \left[ {\mathbf{I} \circ {T_g}} \right]\left( x,y \right),
\end{equation}
where $\mathbf{I}_{rot}$ is a rotated image of $\mathbf{I}$ with a $g^{o}$ angle, and $\mathbf{R}_g$ is a rotation matrix. Generally, the phase function of samples with direction information, e.g., gradient field, SIFT, is a tensor-valued function $D$. Both the coordinate and the tensor values have to rotate, which can be expressed by
\begin{equation}
\begin{array}{l}
\begin{aligned}
{D_{rot}}(x,y): &= {\mathbf{R}_g}D\left( {{T_g}\left( x,y \right)} \right) = {\mathbf{R}_g}D\left( {\mathbf{R}_g^{-1}(x,y)} \right) \\
&= {\mathbf{R}_g}\left[ {D \circ{T_g}} \right]\left( x,y \right).
\end{aligned}
\end{array}
\label{eq2}
\end{equation}
\algdef{SE}[DOWHILE]{Do}{doWhile}{\algorithmicdo}[1]{\algorithmicwhile\ #1}%
\begin{algorithm}[!t]
\footnotesize
  \caption{: FRIFB in the process of training}
  \label{alg::conjugateGradient}
  \begin{algorithmic}[1]
    \Require
      training sample $Tr\_S = \left[ {{x_1}, \cdots , {x_N}} \right]$, FourierHOG parameters $pFou$, boosting parameters $pBoost$.
    \Ensure
      detector $FRIFBDet$
     \Procedure {$RotationInvariantDet$}{$Tr\_S,pBoost$}
     \Do
       \State {$TrainWin = SampleWins(Tr\_S)$}
       \Comment {Fix window size}
       \State {$FeaGen = FourierHOG(TrainWin, pFou)$}
       \Comment {FourierHOG}
       \State {$TrainFea = ACF(FeaGen)$}
       \Comment {ACF}
       \For {t=1:T}
     \State {${h_t,\varepsilon _t} = DT(TrainFea,pBoost)$}
     \Comment {Decision Trees}
       \EndFor
       \State {$FRIFBDet = \sum\limits_{t = 1}^T {{\frac{\varepsilon _t}{1 - {\varepsilon _t}}}}  * {h_t}$}
     \Comment {Boosting}
     \doWhile{${\varepsilon _t} \to 0$}
    \EndProcedure
    \\
      \Procedure {$FourierHOG$}{$TrainWin, pFou$}
      \For {order=1:K}
      \State {$mag,phase = grad(TrainWin)$}
      \State {$f_g = mag. * {e^{ - i * order * phase}}$}
      \State {$FeaGen = RegionConv(f_g)$}
      \EndFor
      \EndProcedure
  \end{algorithmic}
\end{algorithm}

Therefore, as long as the vector rotation degenerates into scalar rotation, rotation invariant feature maps can be obtained. Rotation invariance is analyzed more effectively in polar coordinate where the features can be separated as the angular part and radial part $P(r)$, respectively. An optimal angular information can be represented by such a Fourier basis defined as ${\psi _m}\left( \varphi  \right) = {e^{im\varphi }}$, where $m$ stands for rotation order. In \cite{IJCV2014}, the rotation behaviors $g(\bullet)$ in Fourier domain can be modeled by a multiplication or convolution operator as follows
\begin{equation}
\begin{array}{l}
g\left(\textbf{F}_{m_1} * \textbf{F}_{m_2} \right) = {e^{-i({m_1} + {m_2}){\alpha _g}}}\left[\textbf{F}_{m_1} * \textbf{F}_{m_2} \right] \circ T_{g}\\
g\left(\textbf{F}_{m_1}\textbf{F}_{m_2} \right) = {e^{-i({m_1} + {m_2}){\alpha _g}}}\left[\textbf{F}_{m_1}\textbf{F}_{m_2} \right] \circ T_{g},
\end{array}
\label{eq3}
\end{equation}
where $\{\textbf{F}_{m_i}\}_{i=1}^{2}$ are defined as their Fourier representations in polar coordinate.

A pixel-wise amplitude and phase value, denoted as  $(\textbf{D}(x,y),\theta(\textbf{D}(x,y)))$, is obtained by computing the gradients in the discrete coordinate, which can be seen as a continuous impulse function represented by $h(\varphi):=\|\mathbf{D}(x,y)\|\delta(\varphi-\theta(\mathbf{D}(x,y)))$ \cite{IJCV2014}. Therefore, the Fourier representation of $h(\varphi)$ can be formulated by
\begin{equation}
\textbf{F}_{m}(x,y)  = \frac{1}{2\pi}\int_0^{2\pi} {h\left( \varphi  \right){e^{ - im\varphi }}} = \left\| \textbf{D}(x,y) \right\|
{e^{ - im\theta \left( \mathbf{D}(x,y)\right)}}.
\label{eq4}
\end{equation}

Relying on the shift properties of Fourier transform under polar coordinate, $\textbf{F}_{m}(x,y)$ with a $g^{o}$ relative rotation is defined as
\begin{equation}
{g}\textbf{F}_{m}(x,y) = {e^{im{\alpha _g}}}\left[ {\textbf{F}_{m}}(x,y)\circ {{T}_g} \right].
\label{eq5}
\end{equation}

In order to make the feature rotation-invariant, namely $\textbf{F}_m = g\textbf{F}_m$, we can construct a set of self-steerability (convolution kernels) with the same rotation order by inverse Fourier transformation. According to Eq. (\ref{eq3}), once satisfying $m_1+m_2 = 0$, we can get
\begin{equation}
\begin{array}{l}
g\left(\textbf{F}_{m_1} * \textbf{F}_{m_2} \right) = \left[\textbf{F}_{m_1} * \textbf{F}_{m_2} \right] \circ {T}_{g},
\end{array}
\label{eq6}
\end{equation}

The convolutional features in Eq. (\ref{eq6}) can be seen as the final rotation-invariant representation (please refer to \cite{IJCV2014} for more proof in details.).
\subsection{Aggregate Channel Features (ACF)}
The ACF is simply computed by subsampling the rotation-invariant channel maps with a preset scaling factor, as shown in Fig. \ref{fig1}. This is a pooling-like operation, which has demonstrated its robustness to shifted and rotated deformations to some extent. With the increase of the factor, the features are represented from finely to coarsely, while the feature structure is gradually enhanced.
\subsection{Fast Pyramid Generative Model (FPGM)}
In \cite{Ruderman1994}, Ruderman \textit{et al.} has theoretically proven that the ensemble of scenes (natural images) has statistics which are invariant to scale. Following it, Piotr \textit{et al.}\cite{PAMI2014} extended this theory and proposed a fast image feature pyramid with an application to pedestrian detection. This technique can effectively achieve a feature channel scaling. That is, the features in any image scale ($s_1$) can be directly obtained with a product of a scale-based ratio factor defined by $({s_1}/{s_2})^{-\lambda}$ and the features extracted on a given (known) scale ($s_2$) (More details can be found in \cite{Ruderman1994} and  \cite{PAMI2014}.), which is formulated as
\begin{equation}
\textbf{F}_{s_1} \approx \textbf{F}_{s_2} \cdot ({s_1}/{s_2})^{-\lambda},
\label{eq7}
\end{equation}
where $\{\textbf{F}_{s_i}\}_{i=1}^{2}$ denote the rotation-invariant feature maps extracted from the different image scales \cite{PAMI2014}. Given $\mathbf{F}_{s_1}$ and $\mathbf{F}_{s_2}$ and the corresponding scale ratio ($s_1/s_2$), the scaling factor $\lambda$ is simply estimated by Eq. (\ref{eq7}) before training and testing model.

The FPGM can compute finely sampled feature pyramids by feature scaling with octave-spaced scale intervals without losing performances. Nevertheless, the input in FPGM needs to meet a low-level feature invariance, hence the Fourier-based rotation invariant feature maps can be perfectly embedded into this framework to correct the bias and variance of the trained classifier caused by various deformations (e.g., rotation, shift).
\section{Experiments}
\subsection{Data Description and Experiment Setup}
The NWPU VHR-10 dataset \cite{Cheng2014} is used as the benchmark data for assessing the performances of the mentioned algorithms. It is collected from Google Earth with a spatial resolution of 0.5m to 2m and infrared images with a 0.08m spatial resolution obtained from the Vaihingen data set provided by the German Society for Photogrammetry, Remote Sensing and Geoinformation (DGPF). We selected two scenes including airplanes and baseball diamonds from this datasets for deeply analyzing and discussing the superiority and effectiveness of the proposed FRIFB. In our experiments, 60\% scene images are randomly selected for training, and the rest for test. Moreover, the positive samples in the training set is simply augmented by mirror processing, while the negative ones are randomly selected from 150 images without any targets. The average size of airplanes and baseball diamonds are 75 $\times$ 75 and 90 $\times$ 90, respectively. For a fair comparison, we conducted 5-fold cross-validation and report an average result.

Similarly to classical object detection methods, we employ the same indices, precision recall curve (PRC) and average precision (AP), to quantitatively evaluate the performances in geospatial object detection. More precisely, if the intersection over union (IoU) ratio between the detection bounding box and the ground-truth box exceeds 0.5, then it is counted as a true positive (TP); otherwise, as a false negatives (FN).
\begin{figure}[!t]
	  \centering
		\subfigure[Baseball Diamond]{
			\includegraphics[width=0.44\linewidth]{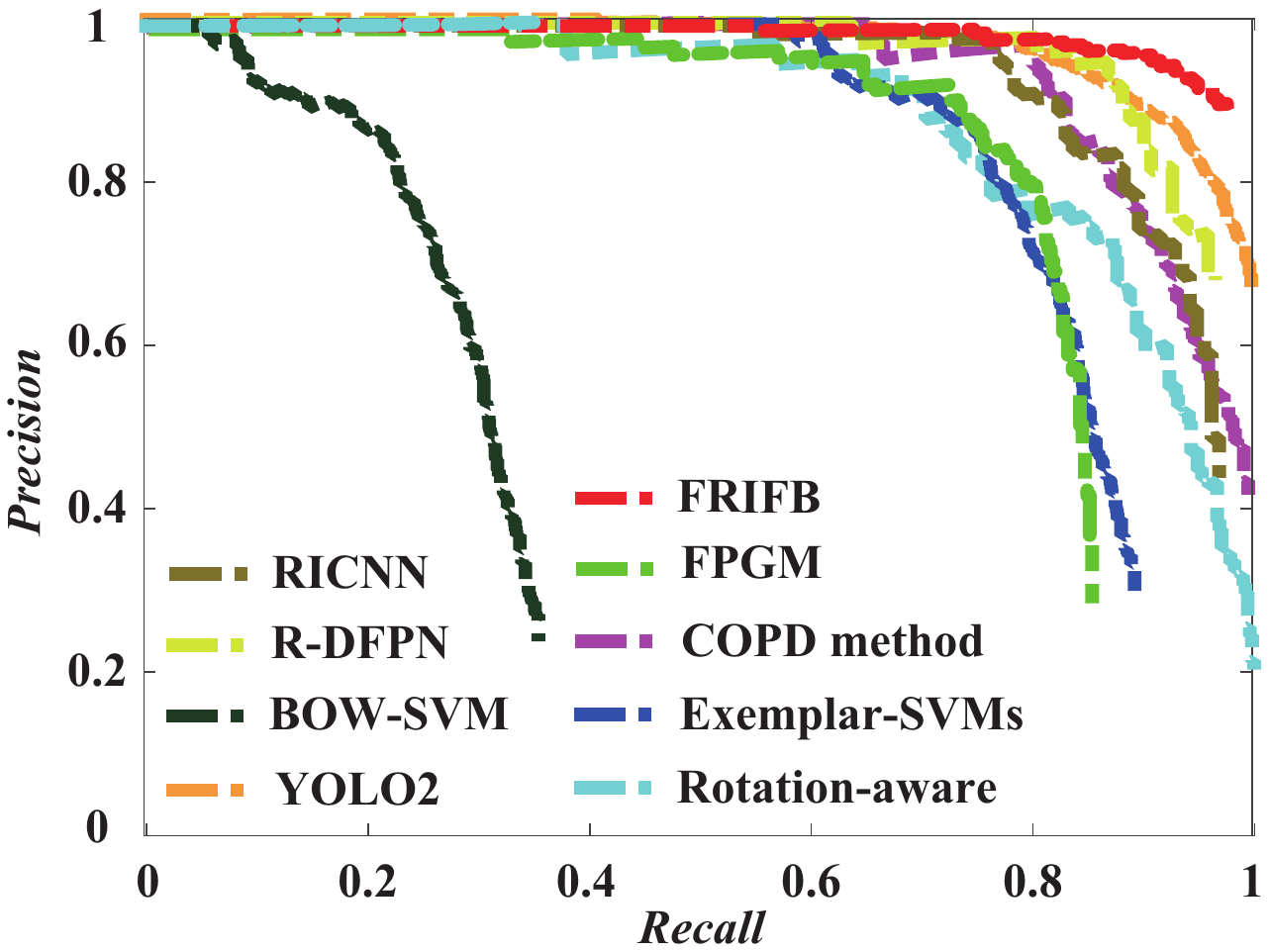}
		}
		\subfigure[Airplane]{
			\includegraphics[width=0.44\linewidth]{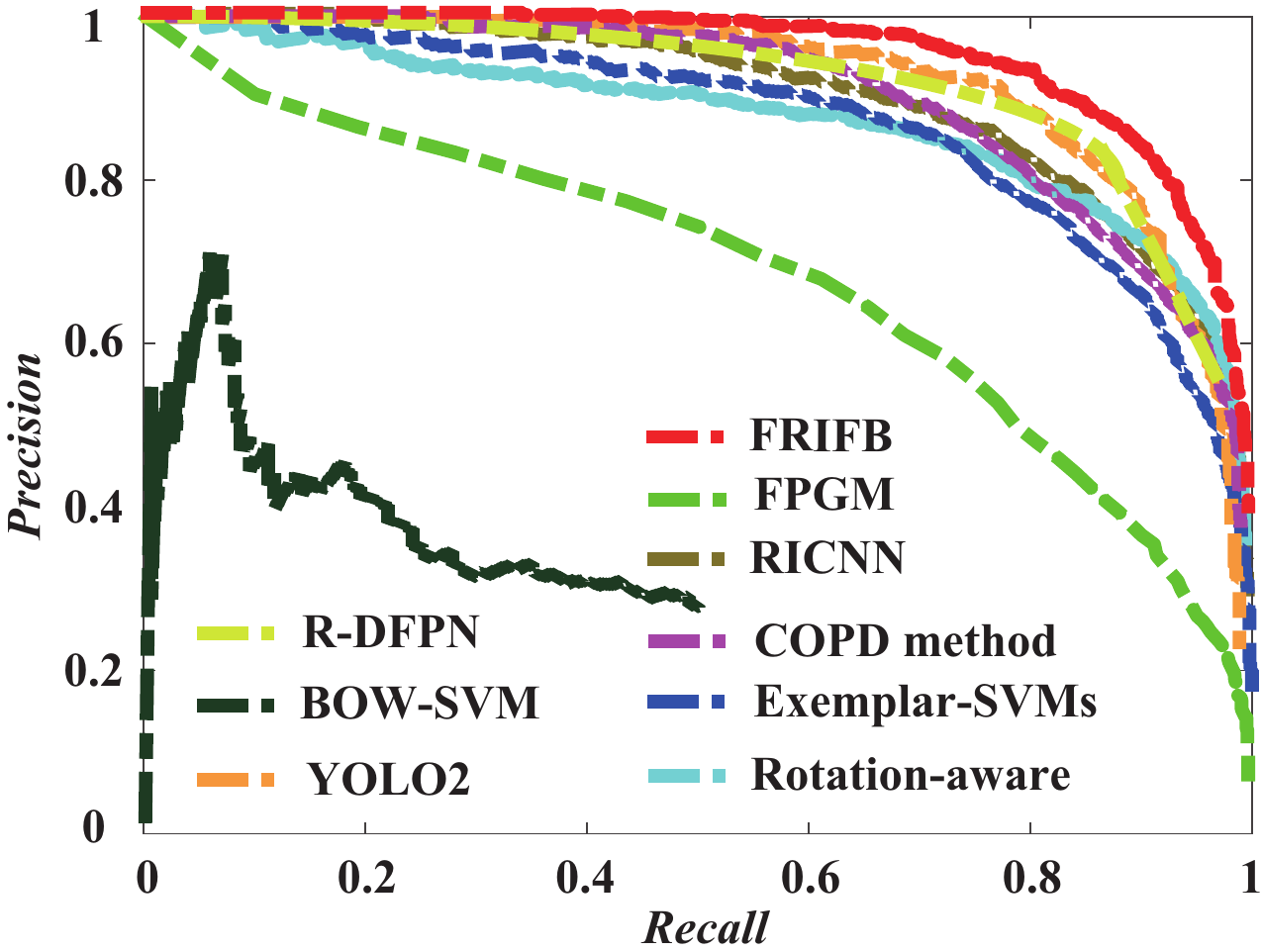}
		}

         \caption{Precision recall curves (PRC) of the proposed framework and some state-of-the-art approaches for airplane and baseball diamond, respectively.}
\label{fig3}
\end{figure}
\begin{table*}[!t]
 \centering
 \caption{Performance comparisons of different methods in terms of AP values and average running time per image. The best results are shown in bold.}
     \begin{tabular}{cccccccccc}
         \toprule[1.5pt]
         Methods & BOW-SVM & Rotation-aware & Exemplar-SVMs& COPD& R-DFPN & RICNN & YOLO2 (GPU) & FPGM & FRIFB \\
         \hline
         Airplane & 25.12 & 79.35 & 80.37 & 85.13 & 88.60 & 86.64 & 89.75 & 65.59 & \textbf{90.20} \\
         \hline
         Baseball diamond & 35.28 & 84.55 & 79.16 & 88.93 & 90.80 & 88.23 & 94.32 & 79.17 & \textbf{97.34} \\
         \hline
         Mean times (s) & 0.85 &1.63 & 1.49 & 0.97 & 0.7 & 8.77 &\textbf{0.13}& 0.16 & 0.88 \\
         \bottomrule[1.5pt]
     \end{tabular}
 \label{tab1}
\end{table*}
\subsection{Detection on NWPU VHR-10 dataset}
\begin{figure*}[!t]
\centering\includegraphics[width=0.9\linewidth]{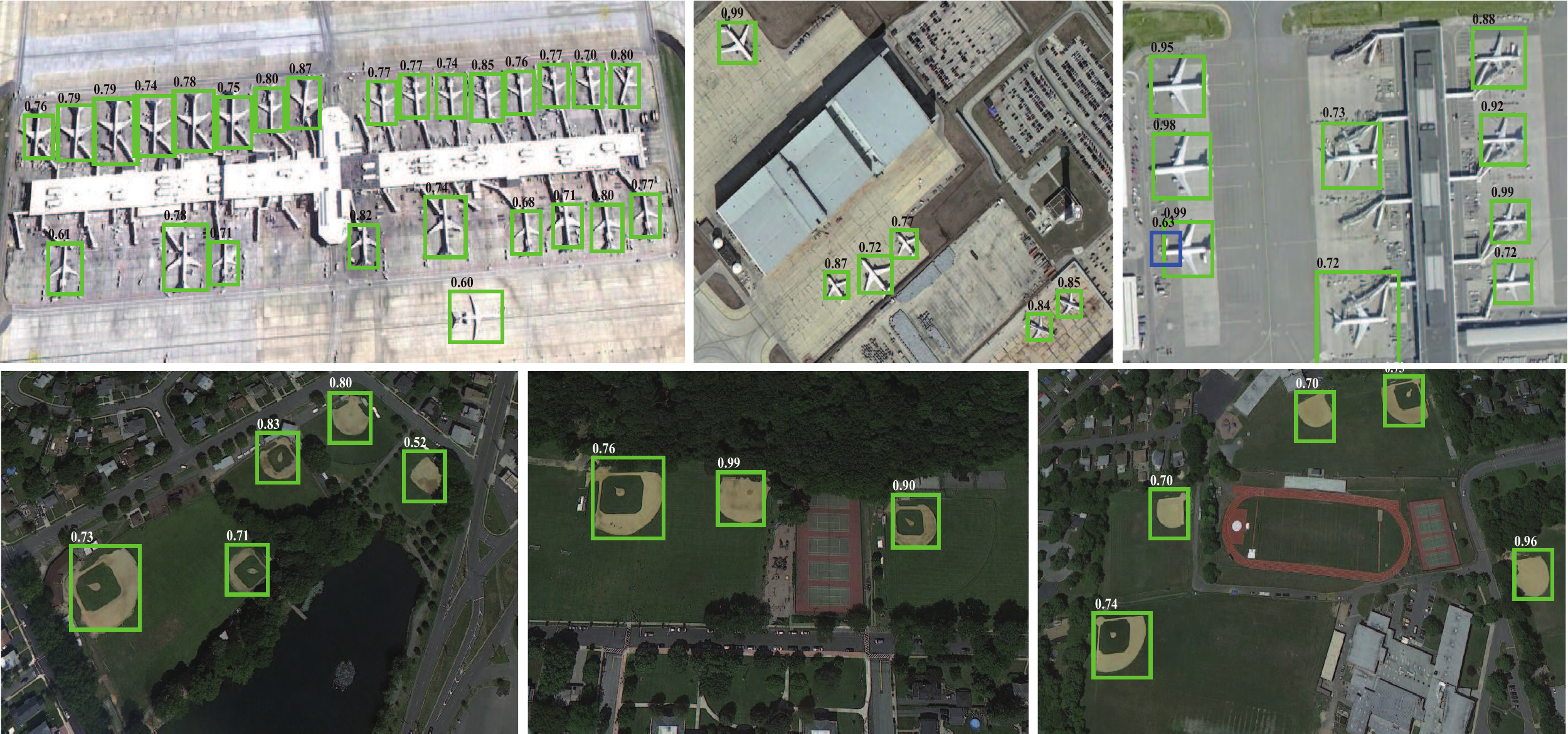}
\caption{Airplane and Baseball diamond detection results with the proposed approach on NWPU VHR-10 datasets.}
\label{fig2}
\end{figure*}

Fig. \ref{fig2} shows the visual performance of the detection results using FRIFB, where the green and blue boxes indicate the correct localization and the false alarm, respectively. As expected, the proposed method detects most of targets with less false positive results, demonstrating the robustness and effectiveness various rotation behaviors. However, the feature discrimination still remains limited, particularly when detecting the airplane's tail fin and the edges or corners of the ground track field. Note that the robustness of the proposed descriptor mainly lies in the resolution of training samples and the Fourier basis functions setting. That means that as long as the training samples and the basis functions can be sufficiently sampled, then the robustness against object rotations and complex noises can be theoretically guaranteed.

\subsection{Comparison with State-of-the-Art Algorithms}
To effectively evaluate the performances of the proposed method (FRIFB), we make a comparison with some state-of-the-art algorithms: BOW-SVM \cite{BOW2010}, rotation-aware features \cite{Schmidt2012}, exemplar-SVMs \cite{Malisiewicz2011}, COPD \cite{Cheng2014}, R-DFPN \cite{yang2018automatic}, rotation-invariant CNN (RICNN) \cite{Cheng2016}, you only look once (YOLO2)\footnote{The code we used, including data augmentation, is available from the website: https://github.com/ringringyi/DOTA\_YOLOv2.} \cite{YOLO2017}, FPGM \cite{PAMI2014}. Similarly to RICNN, data augmentation by rotating or translating the training samples with various angles are performed in all compared methods. For the parameter setting of these compared algorithms, please refer to the corresponding references for more details.

We visually observe the trends of PRC and AP values for the seven different methods in the two different scenes, as shown in Fig. \ref{fig3}. Correspondingly, Table \ref{tab1} lists quantitative comparison results in terms of AP values and running time per image. More specifically, BOW-SVM yields poor performances, since it ignores to model the spatial contextual relationships, leading to a low-discriminative feature representation. Considering the rotation behavior of the objects, Exemplar-SVMs and Rotation-aware methods perform better, but the features constructed in a discrete grid still hinders their performances. Besides, the computational costs for the above methods are expensive, in particular computing image pyramid features. The detection method based fast feature pyramid is effectively run in real-time. Furthermore, fast feature pyramid consists of 10 channels, such as LUV color channels (three channels), normalized gradient magnitude (one channel), and histogram of oriented gradients (six channels), which can achieve better performances in baseball diamond samples and faster running speed. Owing to the well-designed network architecture and GPU's high performance computing, YOLO2 achieves a fastest running speed with a competitive detection accuracy. Nevertheless, the detection performance of YOLO2 is still inferior to that of the proposed FRIFB at around $1\%$ and $5\%$ levels on the two datasets, respectively, as the YOLO2's sensitivity to those tiny and arbitrary pairs of objects hurts its performance to some extent. Not unexpectedly, the proposed FRIFB outperforms others in terms of precision. Without relying on the sample augmentation in the training process, FourierHOG focuses more on designing the intrinsic rotation property by simultaneously considering the local and global information of the image. Please note that although the YOLO-like methods hold a lower computational cost, yet for many applications, such as precision agriculture, urban planning that needs to accurately collect the building information, they prefer to pursue the higher detection accuracy with an acceptable running time. As a result, our proposed FRIFB might be applicable to some practical cases.

\section{Conclusion}
In this letter, we revisit the fast pyramid feature method which is sensitive to rotation and provide an effective remedy by introducing a rotation-invariant descriptor. This descriptor is tightly integrated into the power law, which can fundamentally correct the bias and variance of the trained classifier caused by rotation. Furthermore, we develop a novel and efficient framework for geospatial object detection framework by integrating multi-techniques that have complementary advantages. Extensive experimental results indicate the proposed method is robust to rotation and can effectively improve the detection performance. In the future work, we will focus on tiny object detection by developing an end-to-end learning framework (e.g., deep learning) or introducing auxiliary data (e.g., hyperspectral or multispectral data \cite{hong2019augmented}).


\ifCLASSOPTIONcaptionsoff
  \newpage
\fi

\bibliographystyle{IEEEbib}
\bibliography{reference}
\end{document}